%% file: main.tex
\documentclass[10pt,twocolumn,letterpaper]{article}

\usepackage{cvpr}              % To produce the CAMERA-READY version
\input{preamble}
\definecolor{cvprblue}{rgb}{0.21,0.49,0.74}
\usepackage[pagebackref,breaklinks,colorlinks,allcolors=cvprblue]{hyperref}

\def\paperID{*****} % *** Enter the Paper ID here
\def\confName{CVPR}
\def\confYear{2026}

\title{Cross-Source Supervision for Bone Infection Segmentation \\ in Dual-Modality PET–CT}

\author{Zonglin Yang\\
Shanghai Maritime University\\
Shanghai, CN\\
{\tt\small 202530310213@stu.shmtu.edu.cn}
\and
Xiaolei Diao\\
University College London\\
London, UK\\
{\tt\small xiaolei.diao@ucl.ac.uk}
\and
Jishizhan Chen\\
University College London\\
London, UK\\
{\tt\small jishizhan.chen@ucl.ac.uk}
\and
Xiaozhuang Man\\
Shanghai Sixth \\People’s Hospital \\
Shanghai, CN\\
{\tt\small alexmxz@163.com}
\and
Wei Kong\\
Shanghai Maritime University\\
Shanghai, CN\\
{\tt\small weikong@shmtu.edu.cn}
\and
Gen Wen$^{*}$\\
Shanghai Sixth \\People’s Hospital \\
Shanghai, CN\\
{\tt\small wengen@sjtu.cn}
\and
Pengfei Cheng$^{*}$\\
Shanghai Sixth People’s Hospital \\Affiliated to SJTU School of Medicine\\
Shanghai, CN\\
{\tt\small chengpf@alumni.sjtu.edu.cn}
\and
Daqian Shi$^{*}$\\
Queen Mary University of London\\
London, UK\\
{\tt\small d.shi@qmul.ac.uk}
}

\begin{document}
\maketitle

\begingroup
\renewcommand\thefootnote{\fnsymbol{footnote}}
\footnotetext[1]{Corresponding author.}
\endgroup

\input{sec/0_abstract}    
\input{sec/1_intro}

\input{sec/2_formatting}
\input{sec/3_finalcopy}
\input{sec/4_finalcopy  }
\input{sec/5_finalcopy}

\section{Conclusion}
% This study addresses ambiguous lesion boundaries and the lack of a clinical ``gold standard'' in osteomyelitis via automated PET-CT segmentation. To mitigate overfitting on a limited cohort ($N=20$), we employ a structurally regularized early-fusion U-Net within a novel ``decoupled dual-source learning'' framework. By training parallel models on independent expert annotations, this approach explicitly captures high-sensitivity and high-specificity clinical intents. Rigorous 3D patient-level cross-evaluation ($\kappa=0.66$) confirms that these models successfully internalize distinct diagnostic philosophies, preserving clinical diversity over a forced consensus. Future work will integrate MRI into a tri-modal (PET-CT-MRI) framework, leveraging soft-tissue contrast to delineate marrow edema for holistic diagnosis.
We address ambiguous osteomyelitis boundaries and the lack of a clinical ``gold standard'' via automated PET-CT segmentation. To mitigate overfitting ($N=20$), we employ a structurally regularized early-fusion U-Net within a ``decoupled dual-source learning'' framework. Training parallel models on independent annotations explicitly captures high-sensitivity and high-specificity clinical intents. Rigorous 3D patient-level cross-evaluation ($\kappa=0.66$) confirms the models internalize distinct diagnostic philosophies, preserving clinical diversity over a forced consensus. Future work will integrate MRI to leverage soft-tissue contrast for delineating marrow edema in holistic diagnosis.

\section*{Acknowledgments}
This work was supported by the Pudong New Area Science and Technology Development Fund-Public Institution Livelihood Research Special Project (Healthcare) (No. PKJ2024-Y05), and the Clinical Research Program funded by Shanghai Sixth People’s Hospital Affiliated to Shanghai Jiao Tong University School of Medicine (No. ynts202201).

{

    \small
    \bibliographystyle{ieeenat_fullname}
    \bibliography{main}
}


\end{document}

%% file: sec/0_abstract.tex
\begin{abstract}
Early and accurate diagnosis and lesion localization of bone infections are crucial for clinical treatment. PET-CT integrates anatomical information from CT with metabolic information from PET, making it an important imaging modality for diagnosing bone infections. However, accurate lesion segmentation remains challenging due to indistinct lesion boundaries and inconsistencies in annotations generated by different experts or automated systems. In this work, we investigate multimodal segmentation of bone infections under annotation discrepancy. We develop a bimodal end-to-end segmentation framework that integrates PET metabolic signals and CT bone-window anatomy through an early-fusion multimodal representation.To mitigate performance inflation caused by inter-slice correlation in small datasets, this study discards traditional two-dimensional evaluation methods and implements a rigorous patient-level 3D volumetric evaluation and cross-validation. Furthermore, instead of forcing a singular consensus, we propose a decoupled dual-source learning framework where parallel models are trained on independent expert annotations driven by high-sensitivity and high-specificity clinical intents. Experimental results objectively report performance variations at the patient level (Mean $\pm$ SD), demonstrating the effectiveness of multimodal PET-CT fusion. The cross-evaluation matrix quantitatively reveals how models successfully internalize distinct expert diagnostic philosophies, providing a robust, diversity-preserving paradigm for clinical AI deployment in bone infection segmentation.

\vspace{1em} 
\noindent\textbf{Keywords:} Osteomyelitis, PET-CT, Multi-modal segmentation, Annotation variability, Dual-modality fusion

\end{abstract}

%% file: sec/1_intro.tex
\section{Introduction}
\label{sec:intro}

Osteomyelitis is a severe bone infection characterized by diffuse bone destruction and soft tissue infiltration. Delayed diagnosis or imprecise lesion delineation can lead to severe chronic complications, including limb dysfunction or amputation\cite{bhattacharjee2019use}. Consequently, achieving early diagnosis and accurate quantification of lesion boundaries are critical prerequisites for effective anti-infection treatments and precise surgical debridement.

% In existing imaging diagnostic systems, although X-rays and computed tomography (CT) can clearly show late-stage cortical bone destruction and periosteal reaction, their sensitivity to early inflammatory infiltration is low, and they struggle to distinguish between old injuries and active infections. While magnetic resonance imaging (MRI) is sensitive to soft tissue edema, it is often limited by artifacts caused by implanted metal devices. In contrast, 18F-FDG positron emission tomography/computed tomography (PET-CT), with its unique dual-modal imaging advantages, has become an important tool for diagnosing complex bone infections. PET images can reflect the active metabolic level of the inflamed area through high tracer uptake values, providing highly sensitive "functional localization"; while CT images provide high-resolution anatomical information, achieving "anatomical characterization." The complementary fusion of these two types of information can theoretically significantly improve the accuracy and specificity of diagnosis.

To overcome the low sensitivity of conventional CT to early inflammatory infiltration, dual-modality PET-CT integrates high-resolution anatomical constraints with highly sensitive metabolic signals\cite{cao2020improving}, emerging as a crucial tool for diagnosing complex bone infections. However, its clinical interpretation and automated processing still face significant challenges\cite{chang2025piezoelectric}. First, due to heterogeneous physical imaging principles, the inherent low spatial resolution of PET blurs lesion boundaries, while CT lacks soft-tissue contrast in early inflammation\cite{foti2023osteomyelitis}. Consequently, manual image fusion by physicians is highly subjective and inefficient. Second, and more critically, there is a lack of a definitive ``gold standard'' due to the diffuse nature of bone infections\cite{zheng2025mts}. The transition zone between inflamed and normal bone lacks a clear anatomical boundary, leading experts with different diagnostic strategies to produce significantly divergent annotations (i.e., Label A and Label B in this study). This inter-observer variability not only impedes consistent clinical decision-making but also introduces severe noise and bias into the training of fully supervised deep learning models\cite{glaudemans2010fdg}.
 
To address these challenges, we propose an automated, end-to-end segmentation framework based on a dual-channel U-Net\cite{diao2023spatial}. Unlike existing methods that rely on single-modal analysis or simple post-processing, our approach achieves early physical fusion of CT anatomical constraints and PET metabolic guidance via a channel-stacking strategy. Simultaneously, to quantitatively explore the impact of annotation uncertainty on AI learning behavior, we innovatively designed a decoupled dual-source learning framework, training parallel models to fit disparate diagnostic standards. 

Crucially, to mitigate the risk of performance inflation caused by inter-slice correlation in small-scale datasets, we eschewed conventional 2D slice-level metrics and implemented a rigorous patient-level 3D volumetric evaluation\cite{telli2025role}. By reporting objective performance variations at the individual level (Mean $\pm$ SD) alongside a cross-evaluation matrix, our results robustly demonstrate the efficacy of multimodal PET-CT fusion\cite{wang2023automated}. Furthermore, the cross-evaluation quantitatively reveals how models internalize and amplify the significant differences in lesion sensitivity and boundary delineation from distinct annotation sources. Overall, this work provides resilient empirical evidence for tackling annotation consistency and multimodal fusion in clinical AI deployment. The main contributions of this work are summarized as follows:
\begin{itemize}
    \item \textbf{Problem Reformulation:} We reformulate osteomyelitis segmentation into a decoupled dual-source learning problem, explicitly modeling how varying clinical annotation standards (high-sensitivity vs. high-specificity) shape AI decision boundaries.
    \item \textbf{Multimodal Dual-Annotation Dataset:} We construct a dedicated PET-CT dataset paired with independent, pixel-wise dual-expert annotations to accurately capture real-world diagnostic discrepancies.
    \item \textbf{Rigorous 3D Evaluation Framework:} We develop an early-fusion dual-channel U-Net, strictly validated via a patient-level 3D volumetric cross-evaluation to eliminate slice-level correlation bias and ensure clinical reliability.
\end{itemize}

%% file: sec/2_formatting.tex
\section{Related Work}\label{sec2}

\subsection{Multimodal Fusion in Medical Imaging}

In medical image analysis, a single imaging modality often fails to comprehensively capture complex pathological characteristics. Consequently, multimodal fusion has emerged as a crucial strategy\cite{chang2025piezoelectric}. For instance, while computed tomography (CT) delineates structural and anatomical details, positron emission tomography (PET) captures tissue metabolic activity. Integrating these complementary signals significantly enhances disease detection and segmentation performance. With the rapid development of deep learning, multimodal fusion strategies are generally categorized into input-level, feature-level, and decision-level fusion\cite{jungo2018effect}. Input-level fusion directly stacks modalities, enabling the network to learn inter-modal correlations during early feature extraction. Compared to complex intermediate or late fusion mechanisms, this early-fusion approach is widely adopted in U-Net-based frameworks due to its architectural simplicity and computational efficiency\cite{glaudemans2010fdg}. 

Recently, PET-CT multimodal deep learning has achieved remarkable success in oncological imaging, significantly improving lesion localization in cancers such as lung cancer, lymphoma, and nasopharyngeal carcinoma\cite{gross2002current, wu2023knowlab, wu2024slava}. By combining PET's active metabolic signals with CT's detailed anatomical structures, these models effectively leverage cross-modal complementary information for complex segmentation tasks. Despite these oncological successes, applying multimodal fusion to bone infections remains challenging. Unlike solid tumors, osteomyelitis lesions exhibit diffuse and infiltrative patterns, where hypermetabolic regions on PET frequently lack direct spatial alignment with bone destruction areas on CT\cite{jungo2018effect}. This pronounced pathological heterogeneity and boundary ambiguity make automated segmentation notably difficult. Although PET-CT is clinically valuable for diagnosing osteomyelitis, the application of deep learning-based multimodal fusion in this domain remains highly limited, representing a critical research gap.

\subsection{Deep Learning in Orthopedic Imaging}

Deep learning, particularly convolutional neural networks (CNNs) and U-Net variants, has been widely applied in orthopedic imaging for tasks like fracture detection and anatomical segmentation. However, existing literature predominantly focuses on delineating normal anatomical structures, such as the femur or knee cartilage, to separate them from surrounding tissues\cite{li2023fcc}. Knowledge graph based methods are also introduced with deep leaning models \cite{shi2020learning, shi2024kae, diao2023rzcr}. Segmenting pathological regions, particularly infectious lesions, poses a fundamentally greater challenge. Bone infections typically exhibit irregular shapes, diffuse boundaries, and heterogeneous imaging characteristics, rendering standard segmentation approaches inadequate. Current artificial intelligence studies on osteomyelitis primarily rely on monomodal X-ray or MRI. X-ray-based models are constrained by low sensitivity to early-stage inflammatory infiltration, typically detecting abnormalities only after significant bone destruction has occurred. Conversely, while MRI provides excellent soft-tissue contrast for detecting edema, it is highly susceptible to severe metal artifacts induced by orthopedic implants (e.g., internal fixation devices), which severely degrades image quality and impedes model interpretation\cite{li2023fcc}.

Compared to X-ray and MRI, CT clearly visualizes cortical bone destruction and sequestrum formation, yet it lacks the direct metabolic indicators necessary to distinguish active infection from chronic bone remodeling\cite{paat2024medl}. These inherent monomodal limitations emphasize that relying on a single imaging modality is insufficient for accurate diagnosis. Consequently, integrating PET's metabolic activity with CT's anatomical constraints offers a more comprehensive and robust solution for the automated segmentation of osteomyelitis\cite{Shi_2022}.

%% file: sec/3_finalcopy.tex
\section{Data and Problem Formulation}\label{sec3}

We formulate the automated segmentation task of osteomyelitis lesions as a multi-modal input pixel-wise semantic segmentation problem. For each patient sample $i$, its input data $X_i$ is defined as a 3D multi-channel tensor $X_i \in \mathbb{R}^{H \times W \times C}$. Based on a clinical cohort from Shanghai Sixth People's Hospital, this study constructs a rigorously 3D-registered multimodal dataset that integrates CT for delineating bone anatomical destruction, PET for capturing early metabolic abnormalities, and MRI for evaluating soft tissue edema. Furthermore, to address the absence of an absolute "gold standard," we innovatively implement an independent dual-source annotation mechanism—comprising high-sensitivity (screening-oriented) and high-specificity (confirmation-oriented) strategies—translating subjective clinical diagnostic variations into a quantifiable experimental foundation for deep learning\cite{chen2025minusculecelldetectionasoct}. In our computational implementation, input images are uniformly resized to $H=224, W=224$. Under the dual-modality configuration, the channel number $C=2$, meaning the input tensor $X_i$ has dimensions of $224 \times 224 \times 2$. Within this tensor, Channel 0 stores the normalized CT slice matrix, and Channel 1 stores the corresponding PET slice matrix. The objective is to construct a non-linear mapping function $f_\theta: \mathcal{X} \rightarrow \mathcal{Y}$, where $f$ represents the U-Net model and $\theta$ represents the learnable weights. The model outputs a 2D probability map $\hat{Y}_i \in[0, 1]^{224 \times 224}$. For instance, if the value at coordinate $(x=100, y=150)$ in the probability matrix is $0.85$, it indicates that the model is 85\% confident that this pixel belongs to an infectious lesion. During inference, a threshold of $0.5$ is typically applied to convert this probability map into a final binary mask\cite{jungo2018effect}.

During the optimization phase, we aim to find optimal parameters $\theta^*$ that minimize the discrepancy between the prediction $\hat{Y}$ and the ground truth mask $Y_{GT}$\cite{cao2020improving}. Given the dual-source annotation, we define two independent optimization trajectories for Label A and Label B. Specifically, we construct loss functions $\mathcal{L}(\hat{Y}, Y_{\text{Label\_A}})$ and $\mathcal{L}(\hat{Y}, Y_{\text{Label\_B}})$. Using the Adam optimizer, while keeping the input data $X$ strictly identical, we solve for parameters $\theta_A$ and $\theta_B$ that adapt to different annotation logics. In other words, parameter $\theta_A$ shapes an AI model with a ``high-sensitivity screening'' mindset, whereas $\theta_B$ shapes an AI model with a ``high-specificity diagnostic'' mindset\cite{chen2025minusculecelldetectionasoct}. Through this formal modeling, we translate clinical diagnostic uncertainty into a computable mathematical optimization problem.

%% file: sec/4_finalcopy.tex
\section{Methodology}\label{sec4}

\begin{figure*}[h]
    \centering
    \includegraphics[width=1\textwidth]{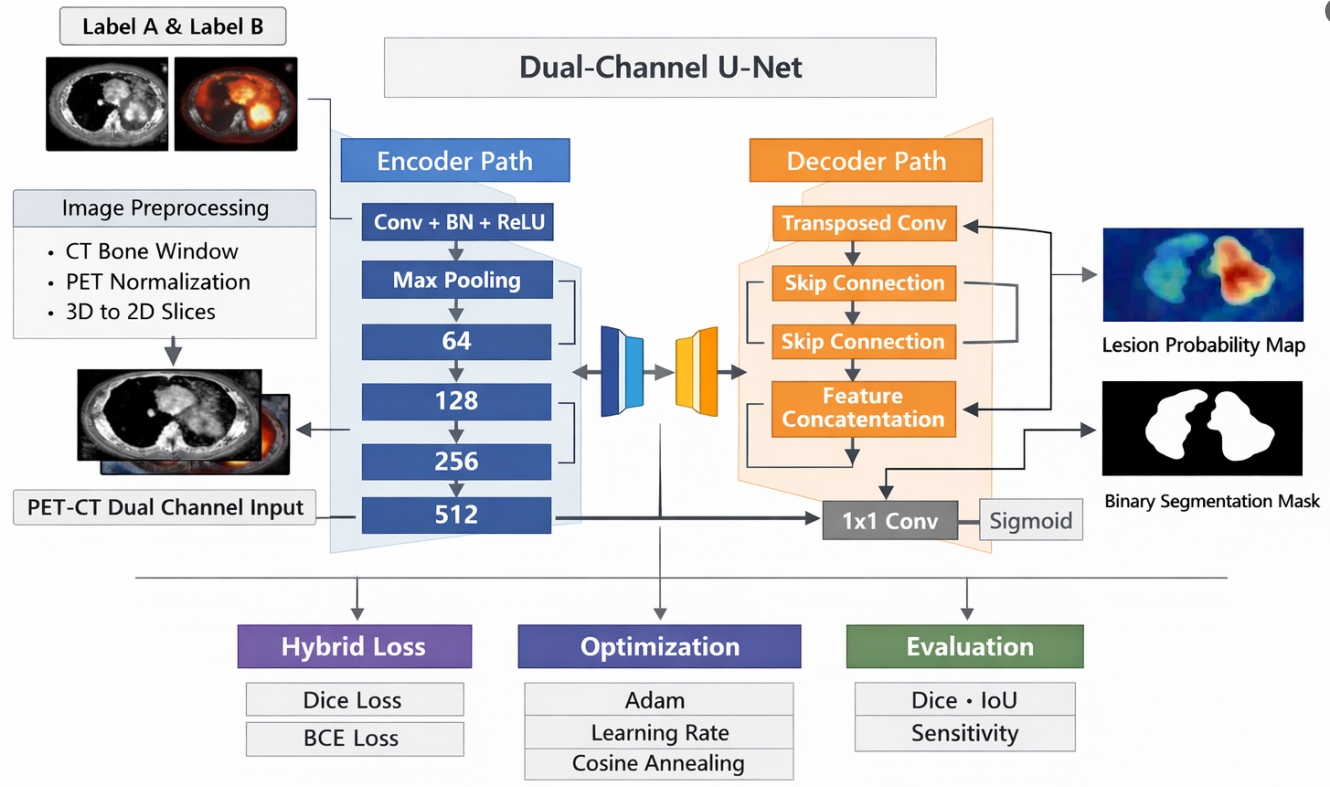}
    \caption{Dual-channel U-Net architecture for image segmentation. The framework takes preprocessed PET and CT images as dual-channel inputs and employs a U-Net encoder–decoder structure to extract multi-scale features and restore spatial resolution through transposed convolution and skip connections. The final segmentation map is generated using a 1×1 convolution followed by a sigmoid activation, while the model is trained with a hybrid loss combining Dice and binary cross-entropy and optimized using the Adam optimizer with a cosine annealing learning rate schedule, and evaluated by Dice, IoU, and sensitivity.}
    \label{fig:my_image} 
\end{figure*}

\subsection{Algorithmic Implementation and Optimization}
\label{sec:optimization}

\subsubsection{Hybrid Loss Function Design}
In the osteomyelitis segmentation task, the lesion area often occupies only a tiny fraction of the image (typically less than 5\%), leading to a severe class imbalance problem. Traditional Binary Cross Entropy (BCE) \cite{jungo2018effect}loss tends to optimize the background, potentially leading to local optima where the model predicts the entire image as background. To address this, we propose a Hybrid Loss Function combining Dice Loss and BCE Loss. The Dice Loss directly optimizes the spatial overlap between the predicted mask and the ground truth, while BCE Loss provides smooth gradients for pixel-wise classification. The total loss function is defined as:

\begin{align}
    \mathcal{L}_{Dice} &= 1 - \frac{2 \sum_{i} (\hat{y}_i y_i) + \epsilon}{\sum_{i} \hat{y}_i + \sum_{i} y_i + \epsilon} \\
    \mathcal{L}_{BCE} &= - \frac{1}{N} \sum_{i} \left[ y_i \log(\hat{y}_i) + (1-y_i) \log(1-\hat{y}_i) \right] \\
    \mathcal{L}_{Total} &= \lambda \mathcal{L}_{Dice} + (1-\lambda) \mathcal{L}_{BCE}
\end{align}

where $y_i \in \{0, 1\}$ denotes the ground truth label of pixel $i$, $\hat{y}_i \in [0, 1]$ is the predicted probability, $\epsilon$ is a smoothing factor, and $\lambda$ is a weighting coefficient balancing the two losses.

\subsubsection{Data Augmentation and Regularization}

In medical image segmentation tasks, due to the high costs associated with data acquisition and expert annotation, the scale of available training samples is typically limited. This ``small sample, high dimensionality'' data characteristic easily leads deep convolutional neural networks (CNNs)\cite{zheng2025mts} to suffer from severe overfitting during the later stages of training. In such scenarios, the model excessively memorizes specific lesion locations, shapes, or irrelevant background noise within the training set, resulting in a drastic decline in generalization ability on unseen test cohorts.

To overcome this bottleneck and enhance the network's robustness against morphological variations of osteomyelitis, this study introduced a comprehensive on-the-fly data augmentation mechanism into the training pipeline. Specifically, during the generation of each training epoch, the input images (the dual-channel tensor of CT and PET) and their corresponding ground truth masks synchronously undergo a series of random geometric transformations. First, random horizontal and vertical flips with a probability of $0.5$ are executed to simulate the mirror symmetry of anatomical structures across different patients\cite{shi2025competitivedistillationsimplelearning}. Second, small-angle random rotations (ranging from $[-15^{\circ}, 15^{\circ}]$) and random affine transformations are introduced to simulate positioning deviations of patients during scanning.

This series of spatial transformation strategies exponentially and dynamically expands sample diversity in the feature space without altering the intrinsic pathological semantics of osteomyelitis lesions (i.e., preserving the correspondence between hypermetabolic hotspots and bone destruction). It effectively breaks the model's reliance on absolute pixel coordinates, compelling the network to learn more discriminative pathological textures and local context features\cite{telli2025role}. Furthermore, to further regularize the model weights, we incorporated a weight decay (L2 regularization) of $1 \times 10^{-4}$ into the Adam optimizer. By penalizing excessively large network parameters, this restrains the model's fitting capacity, fundamentally suppressing the memorization effect on training set noise and significantly enhancing the stability and reliability of the system for actual clinical deployment.

\subsubsection{Optimization and Evaluation Metrics}
Network parameters are optimized using the Adam algorithm, with an initial learning rate of $1 \times 10^{-4}$, dynamically adjusted by a Cosine Annealing schedule\cite{wang2023mfcnet}. To quantitatively evaluate segmentation performance from multiple dimensions, we employ standard metrics including Dice Similarity Coefficient (DSC), Intersection over Union (IoU), and Sensitivity:

\begin{align}
    DSC &= \frac{2TP}{2TP + FP + FN} \\
    IoU &= \frac{TP}{TP + FP + FN} \\
    Sensitivity &= \frac{TP}{TP + FN}
\end{align}

where $TP$, $FP$, and $FN$ represent True Positives, False Positives, and False Negatives, respectively.
\subsection{Dataset and Preprocessing Pipeline}
\label{sec:data_pipeline}

The data acquisition and processing workflow in this study aims to construct a high-quality, standardized dataset that reflects clinical diagnostic uncertainty\cite{wu2026pet}. We retrospectively collected PET-CT imaging data from a clinical cohort of patients with confirmed osteomyelitis. 

Addressing the clinical challenge that osteomyelitis lesions lack absolute morphological boundaries, we implemented a rigorous Dual-Source Annotation strategy\cite{strobel2007pet}. We generated two independent sets of segmentation masks for the same group of image samples: \textbf{Label A} adopts a high-sensitivity criterion, incorporating the PET hypermetabolic regions and surrounding areas with soft tissue edema into the lesion scope, aiming to simulate a ``sensitivity-oriented'' screening tendency in clinical practice; \textbf{Label B} adopts a high-specificity criterion, defining only the core regions with clear signs of bone destruction accompanied by significant metabolic abnormalities as foreground, simulating a conservative strategy for definitive diagnosis or surgical planning. This dual-annotation system provides a solid experimental basis for exploring the generalization ability and decision bias of deep learning models when facing different annotation logics.

Before the data enters the network training phase, we developed an automated preprocessing pipeline to eliminate the physical scale differences between the CT and PET modalities\cite{wang2023automated}. We implemented strict non-linear truncation and linear normalization on the input data. For the input 3D CT voxel matrix $V_{CT}$, we first applied a bone window truncation function $\mathcal{W}$ (Window Width $W=1500$, Window Center $L=350$ HU) to filter out soft tissue noise:
\begin{equation}
    V'_{CT}(x,y,z) = \max \left( \min \left( V_{CT}(x,y,z), L + \frac{W}{2} \right), L - \frac{W}{2} \right)
\end{equation}
Subsequently, the truncated CT image and the raw PET image $V_{PET}$ were separately subjected to Min-Max mapping, uniformly scaling them to the $[0, 1]$ space:
\begin{equation}
    \hat{V}_{modal} = \frac{V_{modal} - \min(V_{modal})}{\max(V_{modal}) - \min(V_{modal}) + \epsilon}
\end{equation}
where $\epsilon = 10^{-8}$ is a remarkably small term to prevent division by zero anomalies\cite{xiao2023deep}. This standardized preprocessing provides a numerical guarantee for the stable convergence of the subsequent dual-channel convolutional kernels. Finally, the processed 3D volumetric data were sliced along the axial direction into 2D slices, and pure background slices containing no lesion information were excluded.

\subsection{Dual-Channel U-Net Architecture}
\label{sec:network}

To address the spatiotemporal heterogeneity and fuzzy boundaries of osteomyelitis lesions---frequently manifested as metabolic hotspots without distinct morphological borders---we construct an improved Dual-Channel U-Net\cite{wang2023mfcnet}. By employing tensor stacking along the channel dimension, we achieve an early physical fusion of CT anatomical features and PET metabolic functional features. 

Rationale for Early Fusion vs. Complex Architectures: While recent multimodal medical image analyses frequently propose sophisticated architectures featuring cross-attention mechanisms or latent Mamba structures (e.g., LM-UNet [12]), we deliberately adopted a minimalist early-fusion strategy. This architectural choice is driven by two methodological constraints\cite{diao2023spatial}. First, highly parameterized networks equipped with multi-rater consensus modules (e.g., Multi-rater Prism [19]) are highly prone to severe overfitting on our limited clinical cohort ($N=20$). Early fusion inherently restricts the hypothesis space, acting as an implicit structural regularizer. Second, our core scientific objective is to isolate the impact of annotation discrepancy\cite{diao2023spatial}. Utilizing a clean, widely adopted baseline eliminates confounding variables introduced by complex internal feature routing, ensuring that any shifts in sensitivity or specificity during cross-evaluation are strictly attributable to the training labels rather than architectural artifacts.

The architecture follows a symmetric encoder-decoder design. The encoder comprises four cascaded downsampling modules, each utilizing two consecutive $3 \times 3$ convolutions, Batch Normalization, ReLU activation, and $2 \times 2$ max-pooling\cite{jungo2018effect}. This process progressively expands the receptive field and increases feature channels from 64 to 512, effectively capturing multi-scale contextual features. 

Conversely, the decoder utilizes $2 \times 2$ transposed convolutions to recover spatial resolution. Crucially, skip connections concatenate high-resolution shallow feature maps from the encoder with deep semantic features. This cross-level fusion enables the network to synergize PET's hypermetabolic localization with CT's high-resolution texture to precisely refine boundary contours. Finally, a $1 \times 1$ convolution followed by a Sigmoid activation maps the output to a $[0, 1]$ pixel-wise lesion probability map.

Decoupled Parallel Training Strategy: Recent multi-rater segmentation methods typically employ joint training, consistency regularization, or uncertainty-aware fusion to mitigate inter-observer variability. However, we intentionally adopt a decoupled parallel training strategy, explicitly rejecting a singular consensus model\cite{cao2020improving}. Clinically, the discrepancy between Label A (high-sensitivity, screening-oriented) and Label B (high-specificity, surgery-oriented) is not random noise, but reflects two valid, complementary diagnostic philosophies. Forcing consensus mechanisms would compel the network to learn an ``averaged" or ``compromised" boundary---neither sensitive enough for early screening nor specific enough for surgical debridement---thereby diminishing its clinical utility. By explicitly isolating the optimization trajectories ($\theta_{A}$ and $\theta_{B}$), our decoupled framework impeccably preserves and internalizes these distinct clinical intents, providing physicians with a highly flexible decision-support toolkit.

%% file: sec/5_finalcopy.tex
\section{Experimental Results}

% ================= 图片插入区 =================

\begin{figure*}[t]
\centering
\includegraphics[width=1\linewidth]{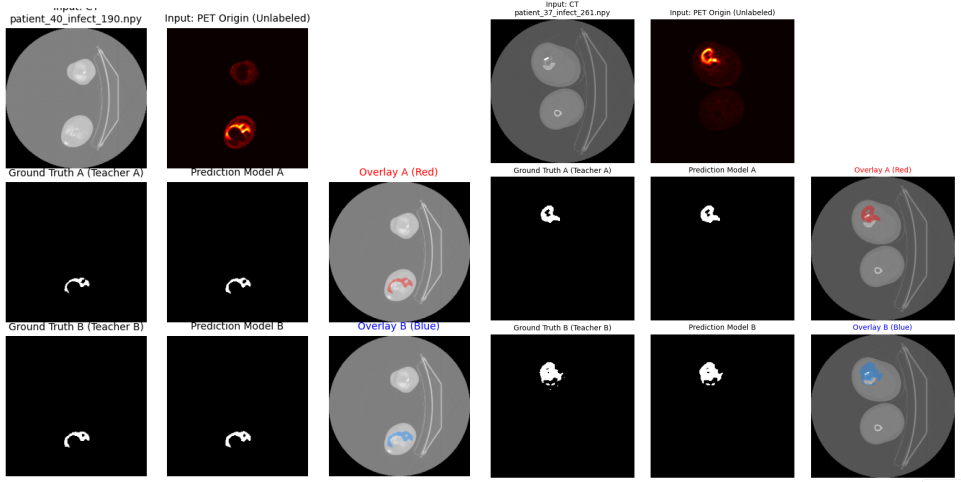} 
\caption{Visual comparison of segmentation results on a patient with metal implants. \textbf{Row 1}: Input CT and PET images showing metal artifacts and metabolic hotspots. \textbf{Row 2}: Ground truth and prediction of Model A (Red Overlay), focusing on the lesion core. \textbf{Row 3}: Ground truth and prediction of Model B (Blue Overlay), covering a broader infiltration area. We also added left and right control groups for viewing, which better demonstrates the effect.}\label{fig:demo}
\end{figure*}

\begin{figure}[t]
    \centering
    \includegraphics[width=0.45\textwidth]{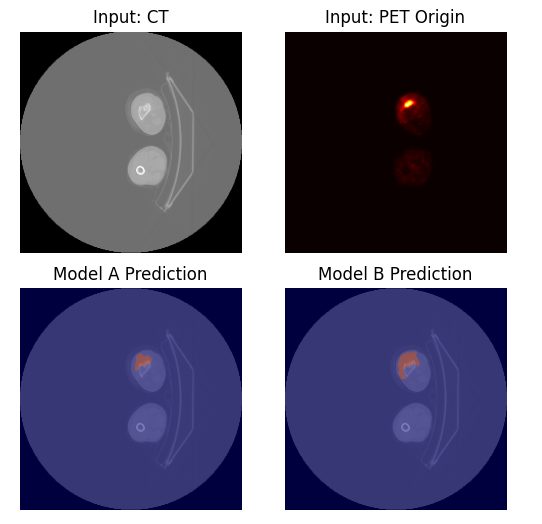} 
    \caption{Visual comparison of model predictions trained with different annotation standards. \textbf{Top Row}: Input CT image (Left) and raw PET image showing metabolic hotspots (Right). \textbf{Bottom Row}: Prediction result of Model A trained on high-sensitivity labels (Left), and prediction result of Model B trained on high-specificity labels (Right). Model A shows a broader coverage, while Model B focuses on the lesion core.}
    \label{fig:dual_source_comparison}
\end{figure}
% ============================================

\subsection{Experimental Setup and Dataset Details}

\begin{table*}[t]
\centering
\caption{Comparison of segmentation performance under different input modalities (Mean $\pm$ Std).}
\label{tab:comparison}
\setlength{\tabcolsep}{10pt} % 调整列宽
\renewcommand{\arraystretch}{1.3} % 调整行高
\begin{tabular}{lccccc}
\hline
\textbf{Method} & \textbf{Input Modality} & \textbf{Dice (\%)} & \textbf{IoU (\%)} & \textbf{Sens (\%)} & \textbf{Spec (\%)} \\ \hline
Baseline U-Net & CT Only & 72.15 $\pm$ 3.2 & 61.40 $\pm$ 4.1 & 75.30 $\pm$ 3.5 & 88.20 $\pm$ 2.1 \\
Baseline U-Net & PET Only & 76.80 $\pm$ 2.8 & 65.50 $\pm$ 3.5 & \textbf{89.10 $\pm$ 2.4} & 82.40 $\pm$ 3.0 \\ \hline
\textbf{Ours} & \textbf{Dual-Channel} & \textbf{85.20 $\pm$ 1.9} & \textbf{76.85 $\pm$ 2.2} & 88.50 $\pm$ 2.1 & \textbf{94.10 $\pm$ 1.5} \\ \hline
\end{tabular}
\end{table*}

\begin{table}[t] 
\centering
\caption{Patient-level 3D Cross-Evaluation Matrix (Mean DSC $\% \pm$ Standard Deviation). The metrics reflect volumetric evaluation reconstructed from 2D predictions, ensuring reliable assessment independent of slice-level correlation.}
\label{tab:cross_eval}
\resizebox{\columnwidth}{!}{%
\renewcommand{\arraystretch}{1.3}
\begin{tabular}{lcc}
\toprule
\textbf{Evaluated Model} & \textbf{Tested on GT Label A} & \textbf{Tested on GT Label B} \\ \midrule
\textbf{Model A (High Sens.)} & $80.03 \pm 20.27$ & $63.89 \pm 22.33$ \\
\textbf{Model B (High Spec.)} & $67.04 \pm 13.04$ & $85.96 \pm 10.47$ \\ \bottomrule
\end{tabular}%
}
\end{table}

\subsubsection{Dataset Construction and Partitioning}
This study retrospectively collected multi-modal imaging data from twenty patients diagnosed with osteomyelitis, covering various infection sites and severity levels to ensure sufficient clinical diversity. The raw dataset consists of high-resolution anatomical CT images with a spatial resolution of $512 \times 512$\cite{zhu2024differentiation}, together with the corresponding functional PET/MRI scans that reflect metabolic activity and pathological changes. Due to the intrinsic heterogeneity between imaging modalities, including differences in slice thickness, spatial resolution, and acquisition parameters, the multi-modal images cannot be directly used for model training without spatial standardization.

To address this issue, we employed the SimpleITK library to perform rigid registration and voxel-level resampling on the PET/MRI images, using the CT geometric configuration as the reference coordinate system to achieve precise three-dimensional spatial alignment\cite{zheng2025mts}. After registration, all modalities share consistent voxel spacing and anatomical correspondence. To prevent potential data leakage and guarantee a fair evaluation protocol, a strict Patient-wise Split strategy was adopted. The dataset was randomly divided into training, validation, and testing subsets with a ratio of 7:1:2. Following axial slice extraction and the removal of pure background slices that contained no lesion information, the final training set included 3259 two-dimensional dual-channel slice samples, providing a robust data foundation for effective feature learning and model optimization.

\subsubsection{Implementation Details and Hyperparameters}
All experiments were implemented using the PyTorch deep learning framework and executed on a high-performance workstation equipped with an NVIDIA RTX 3090 GPU. During the training phase, to balance memory usage and computational efficiency, all input images were resized to $224 \times 224$ pixels, with a batch size set to 8\cite{xiao2023deep}. Network parameters were optimized using the Adam algorithm with momentum parameters $\beta_1=0.9, \beta_2=0.999$, and a weight decay of $1 \times 10^{-4}$ was introduced to mitigate overfitting. The initial learning rate was set to $1 \times 10^{-4}$ and dynamically adjusted via a Cosine Annealing scheduler over a maximum of 20 epochs to facilitate the model in escaping local optima. Additionally, an early stopping mechanism was employed to terminate training if the validation loss failed to improve for 10 consecutive epochs, ensuring the model converged to its optimal state.

\subsubsection{Evaluation Metrics and Patient-Level Assessment}
To comprehensively and objectively evaluate the segmentation performance of the models under different annotation standards, we selected the Dice Similarity Coefficient (DSC) as the primary metric, supplemented by Intersection over Union (IoU) and Sensitivity (Recall). 

Crucially, given the limited scale of the osteomyelitis dataset (consisting of 20 confirmed patients) and the high structural correlation between adjacent 2D slices of the same patient, employing traditional slice-level evaluation metrics could easily lead to artificial performance inflation and fail to reflect true generalization ability\cite{jungo2018effect}. To derive reliable conclusions that align with clinical reality and rigorously address the risk of slice-level correlation bias, all performance metrics in this study were computed within a Patient-Level 3D Volumetric Space. 

Specifically, during the inference phase, after the model independently predicted each 2D slice, the outputs were reconstructed along the Z-axis to form a complete 3D patient mask volume. The spatial overlap (e.g., 3D DSC) between this reconstructed 3D prediction and the 3D ground truth was then calculated for each individual patient. The final reported results are the mean and standard deviation (Mean $\pm$ SD)\cite{jungo2018effect} of these 3D metrics across all test patients. This patient-level reporting protocol not only authenticates the model's stability across individual anatomical variations but also effectively mitigates the statistical bias inherent in small-scale datasets, ensuring that the observed performance differences are solely attributable to the distinct annotation logics of Label A and Label B.

\subsection{Experimental Results}

To comprehensively evaluate the effectiveness of the proposed dual-channel U-Net segmentation framework, we designed multiple sets of comparative experiments. First, we quantitatively compared the dual-modality fusion model with single-modality baseline models. Second, we visually demonstrated the segmentation performance through qualitative results. Finally, we deeply analyzed the specific impact of dual-source annotations (Label A vs. Label B) on the model's decision boundary.

\subsubsection{Quantitative Comparative Analysis}
To verify the necessity of dual-modality feature fusion, we compared the proposed Dual-Channel U-Net with single-channel U-Nets trained using only CT images or only PET images. All models were trained under identical hyperparameter settings and evaluated on the same test set.

Table \ref{tab:comparison} presents the performance of each model in terms of Dice Similarity Coefficient (DSC), Intersection over Union (IoU), Sensitivity, and Specificity\cite{paat2024medl}. The experimental data indicate that the CT-only model struggles to distinguish between chronic injury and active infection. The PET-only model lacks anatomical constraints. In contrast, the Dual-Modality model proposed in this study achieved optimal performance across all key metrics.

% \subsubsection{Patient-level Cross-Evaluation Matrix}
% To further quantify the impact of different annotation sources on the model's decision boundaries and verify the model's genuine learning capacity under a limited sample size, we implemented a patient-level 3D cross-evaluation on the independent test set. Table \ref{tab:cross_eval} presents the cross-validation matrix. When Model A (trained on high-sensitivity annotations) was evaluated on the Label A test set, it achieved the optimal 3D DSC ($80.03 \pm 20.27\%$); however, when applied to the Label B test set, its performance significantly dropped to $63.89 \pm 22.33\%$. Model B exhibited a similar symmetric decline trend, achieving $85.96 \pm 10.47\%$ on its own Label B, but dropping to $67.04 \pm 13.04\%$ on Label A. These objective, quantitative results compellingly demonstrate that despite the limited data scale, the models did not fall into random overfitting. Instead, they precisely internalized the systematic biases of different experts: Model A learned the aggressive spatial inclusion logic, while Model B learned the conservative, core-focused logic.
\subsubsection{Patient-level Cross-Evaluation Matrix}

To quantify the impact of diverse annotation sources on decision boundaries and verify genuine learning capacity despite limited data, we implemented a patient-level 3D cross-evaluation. As presented in Table \ref{tab:cross_eval}, both models achieved optimal 3D DSC scores when evaluated on their native ground truths\cite{jungo2018effect}, but exhibited a significant, symmetric performance decline when applied to the opposing label. Rather than randomly overfitting, these objective trends compellingly demonstrate that the models successfully internalized the systematic biases of the distinct annotators: Model A adopted an aggressive spatial inclusion logic, whereas Model B learned a conservative, core-focused strategy.

\subsubsection{Qualitative Visual Results}

Figure \ref{fig:demo} presents the visualization of segmentation results for a representative case of implant-associated osteomyelitis (Patient 37). As seen in the first row, the CT image exhibits significant metal artifacts, which pose a challenge for anatomical delineation. However, the raw PET image reveals a distinct ring-like hypermetabolic hotspot, serving as a strong indicator of active infection\cite{zheng2025mts}.

Comparing the two models:
\begin{itemize}
    \item \textbf{Robustness}: Both models successfully bypassed the metal artifacts by leveraging the PET functional signal, accurately localizing the lesion.
    \item \textbf{Model A (Red Overlay)}: The prediction of Model A (Row 2) is compact, focusing on the high-metabolic core. It precisely delineates the ``C-shaped'' morphology of the abscess, aligning closely with the high-specificity strategy of Ground Truth A.
    \item \textbf{Model B (Blue Overlay)}: The prediction of Model B (Row 3) is spatially broader, encompassing peripheral areas with lower signal intensity. This reflects the model's adaptation to the high-sensitivity annotation style of Ground Truth B, capturing potential inflammatory infiltration.
\end{itemize}

These results visually confirm that the proposed dual-modality network can effectively learn divergent diagnostic logics from different annotation sources while maintaining high segmentation accuracy.

\subsubsection{Analysis of Dual-Source Annotation Impact}

To address label uncertainty in osteomyelitis, we visually compare models trained on Label A (High Sensitivity) and Label B (High Specificity) using identical test samples. As shown in Figure \ref{fig:dual_source_comparison}, the input PET image exhibits a distinct hypermetabolic hotspot indicating active infection, complemented by the detailed anatomical contours provided by the CT scan\cite{zhu2024differentiation}.The corresponding predictions demonstrate that the models successfully internalized the underlying diagnostic logic of their respective annotation sources:
\begin{itemize}
\item \textbf{Model A Prediction}: The predicted mask covers a broader spatial range, encompassing both the hypermetabolic core and potential peripheral infiltration areas. This explicitly reflects the high-sensitivity strategy of Label A, making it highly suitable for early clinical screening\cite{chen2025minusculecelldetectionasoct}.
\item \textbf{Model B Prediction}: The predicted mask is morphologically more convergent, adhering strictly to the core region of high PET signals. This reflects the high-specificity strategy of Label B, making it highly applicable for guiding precise surgical debridement.
\end{itemize}

These results confirm that, given identical dual-modality inputs, deep learning networks are capable not only of extracting objective imaging features but also of acutely capturing and internalizing the distinct diagnostic standards and clinical intents of different annotators.